\pdfoutput=1

\documentclass[11pt,a4paper]{article}

\usepackage{acl}

\usepackage{times}
\usepackage{multicol}

\usepackage{latexsym}
\definecolor{ao(english)}{rgb}{0.0, 0.5, 0.0}
\definecolor{aurometalsaurus}{rgb}{0.43, 0.5, 0.5}

\usepackage[T1]{fontenc}
\usepackage{graphicx}
\usepackage{colortbl}
\usepackage{lipsum}
\usepackage{booktabs}
\usepackage{array}
\usepackage{arydshln}
\makeatletter
\newcommand{\thickhline}{%
    \noalign {\ifnum 0=`}\fi \hrule height 1pt
    \futurelet \reserved@a \@xhline
}
\makeatletter
\newcommand{\thinhline}{%
    \noalign {\ifnum 0=`}\fi \hrule height 0.00001pt
    \futurelet \reserved@a \@xhline
}

\usepackage{tabularx}
\usepackage{multirow}
\usepackage{dblfloatfix}
\usepackage{amsmath}
\usepackage{latexsym}
\usepackage{colortbl}

\usepackage{caption}

\DeclareCaptionType{equ}[][]

\usepackage[colorinlistoftodos,prependcaption,textsize=tiny]{todonotes}

\usepackage[utf8]{inputenc}

\usepackage{microtype}
\title{Forecasting COVID-19 Caseloads Using Unsupervised Embedding Clusters of Social Media Posts}

\author{Felix Drinkall*, Stefan Zohren*\dag, Janet B. Pierrehumbert*\ddag \\
    *Department of Engineering Science, University of Oxford \\
    \dag The Alan Turing Institute \\
    \ddag Faculty of Linguistics, University of Oxford \\
    \texttt{felix.drinkall@eng.ox.ac.uk}}

\begin{document}
\maketitle

\begin{abstract}
We present a novel approach incorporating transformer-based language models into infectious disease modelling. Text-derived features are quantified by tracking high-density clusters of sentence-level representations of Reddit posts within specific US states' COVID-19 subreddits. We benchmark these clustered embedding features against features extracted from other high-quality datasets. In a threshold-classification task, we show that they outperform all other feature types at predicting upward trend signals, a significant result for infectious disease modelling in areas where epidemiological data is unreliable. Subsequently, in a time-series forecasting task we fully utilise the predictive power of the caseload and compare the relative strengths of using different supplementary datasets as covariate feature sets in a transformer-based time-series model.
\end{abstract}

\section{Introduction}
\label{introduction}

Many papers have shown that web search data can be used to forecast the spread of infectious diseases \cite{lampos2017enhancing}, \cite{Lampos2021}, \cite{McDonalde2111453118}, \cite{Reinharte2111452118}, \cite{Alruily_2022}. Alongside this literature, social media has been exploited for its predictive potential in several other fields such as quantitative finance \citet{xu-cohen-2018-stock}, \citet{sawhney-etal-2020-deep}, logistics forecasting \citet{Ming_passengerflow} and election forecasting \cite{bermingham-smeaton-2011-using}, \cite{HUBERTY2015992}. Research conjoining these two strands has produced results showing that social media can help predict rises in disease caseloads. \citet{iso-etal-2016-forecasting} and \citet{samaras} both used pre-defined keywords in order to predict outbreaks of influenza; words such as "Influenza", "fever", "headache" were selected a-priori. These papers assume that useful feature sets have no geographical variation and use the same features regardless of the regional social dynamics; they also assume that useful features are limited to words that refer to symptoms. To address these limitations, \citet{DrinkallSwissText2021} set more general and objective inclusion criteria. For each of four US state COVID-19 subreddits, all words over-represented in that US state's COVID-19 subreddit compared to the rest of Reddit were considered to be potential keywords for forecasting. The most informative keywords proved to be highly dependent on the target state, and included many that did not refer to symptoms. However, the paper still relied on static word counts that miss more complex information as the discussion unfolds over time. The present paper extracts more informative features from social media data and, to our knowledge, is the first work to incorporate modern NLP techniques in this setting.

\begin{figure}[t!]
    \centering
    \includegraphics[width=\linewidth]{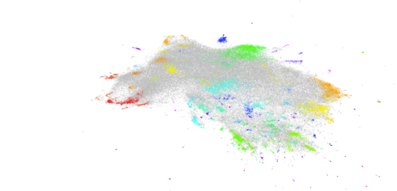}
    \caption{HDBSCAN clusters of the SBERT-NLI-STSb-base representations of r/CoronavirusWA posts made at 50 dimensions but reduced to 2 for visualisation.}
    \label{fig:clustering_hdbscan}
\end{figure}

New transformer-based language models \cite{devlin-etal-2019-bert}, \cite{NEURIPS2019_dc6a7e65_XLNET}, \cite{liu2019roberta} provide the potential for identifying more informative features for infectious disease forecasting, and using them in a more effective manner. This paper uses transfer learning and clustering algorithms to isolate useful features for predicting COVID-19 caseloads. We first pilot-tested a straightforward way to exploit transformer-based language models for the task: the caseload target value was encoded alongside each post in a sequence classification framework. Trained using historical data, this approach generates a prediction from every post, and the results are aggregated for an overall prediction. This method performed very poorly because of noise introduced by irrelevant posts, and we do not discuss it further here (see Appendix \ref{app: ASC_models}). To achieve better performance, we developed a novel feature identification technique that filters out unrelated posts and generates informative features using high-density clusters of posts within a subreddit's embedding space.

Our work builds off \citet{Sia2020} and \citet{thompson2020topic} who demonstrated that clusters of contextualised word embeddings are a good basis for topic modelling. In a similar vein, \citet{Aharoni2020} showed that the domain type of a particular text could be identified using the clustering of sentence-level representations. Finally, \citet{Rother2020} showed that clusters of contextualised embeddings could detect meaning shifts in words. The success of these papers motivates our use of high-density clusters of sentence-level representations.

The present paper shows that our novel feature sets outperform more traditional methods by comparing our results to those in \citet{DrinkallSwissText2021} in a threshold-classification task. This task provides an understanding of which feature sets provide the most informative trend signals at different caseload growth rates, enabling us to understand the effectiveness of a particular feature type at identifying a distinct epidemiological event. Strong performance on this task is relevant in forecasting worst-case scenarios like hospital overflow, where the outcome is a binary variable. 

The caseload information is not fully utilised in the threshold-classification task. This observation motivates a time-series forecasting task to compare feature sets at predicting a more continuous target. Feature selection is a crucial step in time-series modelling \cite{wang2013feature}, \cite{sun2015using}; adding extraneous features to a multivariate prediction can result in performance deterioration as the models get more complex, a fact that inspired L1-regularisation. Only highly relevant features, which represent complementary information, improve performance. 

\textbf{Contributions.} We introduce a novel unsupervised method for predicting COVID-19 trend signals and forecasting caseloads. We show that sole use of our feature set achieves very high accuracy in trend signal prediction, a significant result for infectious disease modelling in regions where other reported data is unavailable or unreliable.

\section{Datasets}
\label{sec: datasets}

Comparing our Reddit features' performance against other open-source geographically-specific datasets allows us to understand their value. The following data sources were used to create the feature sets in this paper:

\textbf{Pushshift API} - The Pushshift API \citep{baumgartner2020pushshift} is used to compile datasets of target subreddits to create the Reddit features. The Pushshift API provides data on every comment and submission posted on Reddit. This paper uses comments to form the subreddit dataset since there are more comments than submissions, and they constitute more conversational and reactionary discourse. No individual comments or users are reported in this paper to observe the anonymity of the users. Update frequency: real-time. 

\textbf{COVID-19 Tracking Project} - The state-level COVID-19 epidemiological data is provided by the {COVID-19 Tracking Project\footnote{\href{https://covidtracking.com}{https://covidtracking.com}}} to create the prediction target and is also used as a feature set in baseline predictions. Update frequency: 24 hours. Start date: 13/01/2020.

\textbf{Oxford COVID-19 Government Response Tracker (OxCGRT)} - The OxCGRT \citep{OxCGRT} defines the local government response. The data covers policies including health, containment and economic measures, and overall stringency scores. Update frequency: "continuously" but can be variable due to human data collection; daily periodicity. Start date: 01/01/2020.

\textbf{Google's COVID-19 Community Mobility Reports (GCCMR) \footnote{\href{https://www.google.com/covid19/mobility/}{https://www.google.com/covid19/mobility/}}} - The GCCMR provides local movement data in different area types such as parks, workplaces, etc. and has been used to successfully predict COVID-19 caseloads \cite{Wang2020.12.13.20248129}, \cite{ilin2021public}. The data is freely available for the duration of the ongoing pandemic. Update frequency: 2-3 days. Start date: 15/02/2020. 

\section{Feature identification}

Social media is a complex and noisy data source, requiring significant processing to isolate meaningful predictive features. The pipeline used in this paper consists of three main steps for feature identification: sentence-level encoding, dimensionality reduction and clustering. This process groups together Reddit comments that are semantically similar. Following these main steps that are outlined below, the Reddit features are reduced further to 25 using a chi-squared test. Once these 25 high-density clusters are identified, the daily counts of comments within these clusters are used as features in the evaluation frameworks in Sections \ref{sec: thresh_class} \& \ref{sec: TS_forc}. 

\subsection{Sentence-level representation}
\label{sec: sent_level_repr}

A common technique for identifying sentence representations is to take the average-pooled BERT hidden-state embedding \cite{Aharoni2020}; however, papers such as \citet{Reimers2019} have shown that the average-pooled BERT embeddings are a relatively poor way of encoding sentences and advocate for further fine-tuning to produce a more semantically meaningful embedding. In \citet{Reimers2019}, the best results are achieved by training the language model on Natural Language Inference (NLI) \citep{bowman-etal-2015-large}, \citep{williams-etal-2018-broad} and Sentence Textual Similarity (STS) \citep{Cer_2017} data. The NLI data contains many sentence pairs with their semantic relationship labelled. The STS data provides a semantic relatedness score between 0-5. It is possible to use both datasets to fine-tune the language model using both dataset types by manipulating the objective functions. The NLI data is trained using a classification objective function, and the STS data is trained using a regression objective function. \citet{Reimers2019} shows that averaging the final layer BERT embeddings leads to a Spearman rank correlation $\rho$ between the cosine similarity of the sentence representations and the actual labels of the STS data of around $\rho = 54.81$, whereas SBERT-NLI-STSb-base achieves $\rho = 88.31$. 

For this paper, there is no domain-specific training. The SBERT-NLI-STSb-base, SRoBERTa-NLI-STSb-base and SDistilBERT-NLI-STSb-base encode the Reddit posts with no further fine-tuning. 

\subsection{Dimensionality reduction}

The language models specified in Section \ref{sec: sent_level_repr} have a dimensionality of 768, which means that their embedding space is very sparse, making it challenging to find dense clusters. Lowering the embedding dimensionality is consistent with the findings in \citet{Sia2020}, who show that the dimensionality of the embeddings can be reduced by $\sim$80\% and still maintain the topic modelling coherence. Therefore, in line with these findings, the dimensionality of the embedding space is reduced to 50.

UMAP (Uniform Manifold Approximation and Projection for Dimension Reduction) is used as in \citet{Rother2020} to lower the dimensionality of the embedding space. UMAP is appropriate for this task since it preserves global structure better than other manifold learning dimensionality reduction methods such as t-SNE \citep{umap} \cite{N2D_McConville}. UMAP's preservation of global structure has been shown in \citet{reif2019_nips} to produce clear clusters related to different word senses. It is tested against a PCA algorithm in Appendix \ref{app: ETM_dim_clust} on the Threshold-Classification task outlined in Section \ref{sec: thresh_class}. The results justify its use as it outperforms PCA when used in conjunction with the best performing clustering algorithm.

\subsection{Clustering}

For this paper, the HDBSCAN algorithm \cite{hdbscan} is used for clustering due to the complex structure of the subreddit embedding space. The benefit of using a density-based clustering algorithm is that sparse areas are not fitted into clusters, removing a significant source of noise from the prediction. 

HDBSCAN offers an advantage over other density-based clustering algorithms; the cut-off density that characterises the edge of the clusters is non-constant and defined by a stability metric that rewards large and dense clusters. This stability metric is calculated from the data points' Minimum Spanning Tree (MST). The following equation defines the stability of cluster $C_i$:

\begin{equation}
    S(C_i) = \sum_{x_j \in C_i} (\lambda_{max}(x_j, C_i)- \lambda_{min}(C_i))
\end{equation}

Here $\lambda$ represents the density statistic: $\lambda = 1/\epsilon$ where $\epsilon$ is equal to the distance between points on the MST. In this equation, $\lambda_{max}(x_j, C_i)$ is the density at which the point $x_j$ would fall out of the cluster $C_i$, and $\lambda_{min}(C_i)$ is the minimum density threshold at which the cluster still exists.  

Clusters with maximum stability are used as the final clusters, and points that fall out of these clusters are discarded. New data points can subsequently be added to the cluster by identifying where they fall in the MST. A point is treated as noise unless it can be grouped into a cluster larger than $min\_cluster\_size$, which, for this paper, we have set at 25 so that the clusters are not too small and the resulting features are not too sparse. Removing noisy comments from the clusters is shown in Appendix \ref{app: ETM_dim_clust} to have performance benefits over other clustering algorithms that do not reject comments: we have compared HDBSCAN to a Spherical K-Means (KM) algorithm and a Gaussian Mixture Model (GMM), two popular algorithms within the literature base.

\section{Threshold-Classification Framework}
\label{sec: thresh_class} 

The threshold-classification framework (henceforth Threshold task) uses the same evaluation methodology as in \citet{DrinkallSwissText2021}. 
The problem is presented as a classification task on balanced classes, with a randomised train/test split and test size of $0.25$ on data from 07/03/2020 to 17/01/2021. Balanced classes allow us to report accuracy as the performance metric for this task. The feature sets, derived from a 7-day moving average of the datasets in Section \ref{sec: datasets}, are concatenated to a target value that encodes whether the caseload increase exceeded the threshold within a given time interval. The threshold is defined by a relative increase, $\delta_r(t)$:

\begin{equation}
    \delta_r(t) = \frac{\mu(t+\tau) - \mu(t)}{\mu(t)}
    \label{eq:rel_thresh}
\end{equation}
\vspace{-0.5cm}
\\

Where $\mu(t)$ is the 7-day moving average of the caseload, and $\tau$ is the prediction horizon.

The model used for classification is a Random Forest (RF) \cite{RF_model}. The advantage of using an RF model over other tree-based models is that it decorrelates the trees, making it robust to correlated feature sets. Social media data is highly correlated as overall take-up surges and wains; therefore, robustness to correlated features is critical. Of course, many more complex models would likely outperform an RF model; however, given that the goal of this task is to compare feature sets, the increased transparency that an RF model offers over more complex models justify its use.

\begin{table}[t!]
    \renewcommand\thetable{1}
    \small
    \centering
    \begin{tabular}{cccccc}
        Feature set & Average & 7D & 14D & 21D & 28D \\
        \hline
        $T_{RoB} ++$ & .875 \cellcolor[gray]{.6} & .895 \cellcolor[gray]{.6} & .880 \cellcolor[gray]{.6} & .849 \cellcolor[gray]{.6} & .874 \cellcolor[gray]{.6} \\
        $T_{BoW} ++$ & .810 & .836 \cellcolor[gray]{.8} & .809 & .805 & .791 \\
        \hline
        $T_{RoB}$ & .803 & .845 \cellcolor[gray]{.8} & .792 & .789 & .787 \\
        $T_{BERT}$ & .789 & .821 \cellcolor[gray]{.8} & .798 & .780 & .761 \\
        $T_{DisB}$ & .780 & .808 \cellcolor[gray]{.8} & .771 & .774 & .768 \\
        $T_{BoW}$ & .768 & .816 \cellcolor[gray]{.8} & .755 & .749 & .753 \\
        $T_{KW}$ & .633 & .733 \cellcolor[gray]{.8} & .628 & .591 & .580 \\
        $M$ & .702 & .703 & .691 & .713 \cellcolor[gray]{.8} & .698 \\
        $G$ & .702 & .713 \cellcolor[gray]{.8} & .710 & .695 & .691 \\
        $P$ & .545 & .516 & .549 & .557 \cellcolor[gray]{.8} & .557 \\
        $C$ & .555 & .651 \cellcolor[gray]{.8} & .536 & .529 & .503 \\
    \end{tabular}
    \caption{The average performance across all relative thresholds and states at different prediction horizons. The features are: $T_{<<language \ model>>}$ \textrightarrow \ our features; $T_{BoW}$ \textrightarrow \ \citet{DrinkallSwissText2021} features; $M$ \textrightarrow \ GCCMR data; $G$ \textrightarrow \ OxCGRT data; $P$ \textrightarrow \ daily post count; $C$ \textrightarrow \ current caseload; $T_{RoB} ++$ \textrightarrow \ $T_{RoB} + M + G + P + C$; $T_{BoW} ++$ \textrightarrow \ $T_{BoW} + M + G + P + C$. The \colorbox[gray]{.8}{light grey} indicates the highest performing instance of each model setup. The \colorbox[gray]{.6}{dark grey} indicates the highest performance for each prediction horizon.}
    \label{tab: time_delay}
\end{table}

\subsection{Evaluation}
\label{sec: TC_evaluation}

Each data type is used in isolation to predict the target labels so that the implicit information within each feature set can be compared. $T_{DisB}$, $T_{BERT}$ and $T_{RoB}$ correspond respectively to the features extracted using the methodology above from the SDistilBERT-NLI-STSb-base, SBERT-NLI-STSb-base and SRoBERTa-NLI-STSb-base language models. The performance of our clustered embedding features is compared against the bag-of-words features used in \citet{DrinkallSwissText2021}, $T_{BoW}$, as well as word-count features taken from a prescriptive list of COVID-19 words defined by the non-hashtag queries in the keyword list in \citet{lamsal}, $T_{KW}$. The evaluation is conducted in four states where Reddit uptake is high: Washington, California, Texas and Florida. The states represent culturally different communities, instilling confidence that the behaviour is true in multiple domains. A successful result across all four states indicates that any observed behaviour is likely not just a symptom of an anomalous community.

The results in Table \ref{tab: time_delay} detail the average performance across the different states and relative thresholds. Firstly, it is clear that word counts from the prescribed list in \citet{lamsal} only capture fractionally more information than a single post-count feature, and that simply using a chi-squared test of over-represented words, $T_{BoW}$, results in a significant performance increase. However, our $T_{DisB}$, $T_{BERT}$, $T_{RoB}$ feature sets perform the best, and when $T_{RoB}$ is used in combination with the comparison feature sets, the performance improves further. It is also evident that as better language models are used, the performance on this task increases. Showcasing the relationship between language model complexity and overall performance supports our a-priori belief that improved semantic information from the text is linked with better epidemiological insights. Due to its success, $T_{RoB}$ alone will henceforth be used in the evaluation as it provides the best performing feature set from our methodology.

\subsubsection{Varying Thresholds}
\label{sec: varying_thresholds}
\begin{table}[t]
    \renewcommand\thetable{2}
    \scriptsize
    \centering
    \begin{tabular}{cccccccc}
        \hline
         & \multicolumn{5}{c}{$\delta_r(t)$} & \\
         \cline{2-7} 
        {$m$} & {$0.2$} & {$0.4$} & {$0.6$} & {$0.8$} & {$1$} & {$\mu + \sigma$}\\ \hline
        $T_{RoB} ++$  & .803 \cellcolor[gray]{.6} & .828 \cellcolor[gray]{.6} & .867 \cellcolor[gray]{.6} & .962 \cellcolor[gray]{.6} & .970 \cellcolor[gray]{.6} & .880 + .039 \cellcolor[gray]{.6}  \\
        $T_{BoW} ++$  & .704 & .795 & .821 & .862 & .910 \cellcolor[gray]{.8} & .809 + .024 \\
        \hline
        $T_{RoB}$  & .753 & .752 & .787 & .876 \cellcolor[gray]{.8} & .876 & .792 + .034 \\
        $T_{BoW}$  & .683 & .699 & .761 & .828 & .865 \cellcolor[gray]{.8} & .755 + .025 \\
        $M$  & .649 & .683 & .663 & .712 & .727 \cellcolor[gray]{.8} & .691 + .026 \\
        $G$  & .678 & .607 & .735 & .761 & .789 \cellcolor[gray]{.8} & .710 + .019 \\        
        $P$  & .548  & .539 & .537 & .530 & .577 \cellcolor[gray]{.8} & .549 + .030 \\ 
        $C$  & .437 & .466 & .527 & .631 & .640 \cellcolor[gray]{.8} & .536 + .039 \\
        \hline
    \end{tabular}
    \caption{Performance across a range of thresholds at a prediction horizon, $\tau = 14$ days, averaged across all states. $\mu$ \& $\sigma$ represent the mean and standard deviation of each feature set's results. The variables and highlighting criteria are the same as Table \ref{tab: time_delay}, but for the \colorbox[gray]{.6}{dark grey} which denotes the highest performance at each threshold.}
    \label{tab: big_table_14}
\end{table}

\begin{table}[b!]
    \scriptsize
    \centering
    \begin{tabular}{ccccccc}
        Feature set & m & Average & 7D & 14D & 21D & 28D \\
        \hline
        \multirow{5}{*}{$T_{RoB} ++$} & $T_{RoB}$ & .331 \cellcolor[gray]{.6} & .334 \cellcolor[gray]{.6} & .307 & .309 & .386 \cellcolor[gray]{.6} \\
         & $M$ & .285 & .264 & .277 & .299 & .301 \cellcolor[gray]{.8}\\
         & $G$ & .307 & .321 & .354 \cellcolor[gray]{.6} & .313 \cellcolor[gray]{.6} & .240 \\
         & $P$ & .009 & .009 & .010 \cellcolor[gray]{.8}& .008 & .009 \\
         & $C$ & .064 & .071 \cellcolor[gray]{.8}& .052 & .070 & .063 \\
         \cline{2-7}
          \multirow{5}{*}{$T_{BoW} ++$} & $T_{BoW}$ & .535 \cellcolor[gray]{.6} & .584 \cellcolor[gray]{.6} & .502 \cellcolor[gray]{.6} & .509 \cellcolor[gray]{.6} & .543 \cellcolor[gray]{.6} \\
         & $M$ & .244 & .222 & .265 \cellcolor[gray]{.8} & .254 & .235 \\
         & $G$ & .171 & .131 & .199 \cellcolor[gray]{.8} & .192 & .162 \\
         & $P$ & .014 & .012 & .009 & .013 & .021 \cellcolor[gray]{.8} \\
         & $C$ & .036 & .050 \cellcolor[gray]{.8} & .025 & .031 & .038 \\
    \end{tabular}
    \caption{Feature importances across varying prediction horizons, at $\delta_r = 0.6$. The variables and highlighting criteria are the same as Table \ref{tab: time_delay}.}
    \label{tab: feature_imp_ETM_time}
\end{table}

Table \ref{tab: big_table_14} breaks down the performance of classifying the data across different threshold increases. Intuitively, the more extreme events are easier to predict, explaining the behaviour across all feature sets. Indeed, when the threshold is large enough, the $T_{RoB} ++$ features achieve an accuracy of .970, significantly higher than the comparison feature sets, showing that social media data is a strong candidate for predicting a sharp rise in caseloads. Again, the performance across all thresholds is highest when using the $T_{RoB} ++$ features as opposed to the $T_{BoW} ++$, highlighting the performance gain from the increased semantic information of transformer-based language models.

\subsubsection{Feature Importance}
\label{sec:feature_imp}

To understand which features more are heavily weighted by the RF model when given the $T_{RoB} ++$ and $T_{BoW} ++$ feature sets, the feature importances are shown in Table \ref{tab: feature_imp_ETM_time}. The tabulated data represents the sum of all individual feature importances in that class. 

Table \ref{tab: feature_imp_ETM_time} shows that despite $T_{RoB}$ performing better than $T_{BoW}$, the other comparison features, $G$ and $M$, are more heavily weighted in $T_{RoB} ++$ than in $T_{BoW} ++$ at some prediction horizons. The $T_{RoB} ++$ feature set performs better than the $T_{BoW} ++$ features, so it appears that the information provided by the $T_{RoB}$ features is complementary to the other feature types. It is also possible that there is some skew in the feature importance owing to the reported over-weighting of more continuous features by a Gini Importance algorithm \cite{strobl}. Regardless of the slight differences, both text-derived feature sets are the most highly weighted when averaged over all prediction horizons, further showing the value of social media in this context.

\section{Time-Series Forecasting Task}
\label{sec: TS_forc}

This section showcases our feature identification methodology within a time-series forecasting framework (henceforth Time-Series task) since this is a widely used prediction task in disease modelling. The high-density clusters are used as covariates in two multivariate time-series models. This setup better utilises the caseload feature and learns the temporal patterns within its historical movement. One difference with the Threshold task in the feature identification pipeline is the feature pruning step that reduces the number of features to 25. In the Threshold task, the target is a binary classification; therefore, a chi-squared test is appropriate. Given that the target is continuous in this task, f-regression is used. F-regression works by firstly calculating the cross-correlation $\rho_i$ of the $i^{th}$ feature $X[:, i]$ and target $y$:

\begin{equation}
    \rho_i = \frac{(X[:, i]-\bar{X}[:, i])\cdot(y-\bar{y})}{\sigma_{X[:, i]}\cdot\sigma_{y}}
\end{equation}

The F-statistic is then calculated along with the associated p-value. Then the top 25 most significant features are filtered to make up the feature set. For each model, the training features and targets are normalised between 0 and 1, and the test set is scaled using the same transformation. The target data is changed from the daily caseload to the daily increase in caseload to make sure the time-series is stationary. No moving average is used since the time-series models should account for the weekly seasonality. The models are trained over 50 epochs on data from 07/03/2020 to 31/12/2020 and tested on data from 01/01/2021 to 01/03/2021. Whilst it is possible to improve the performance by retraining the model on recently evaluated data and sliding the train-test split across the dataset, our proposed framework highlights how the models perform on completely out-of-sample data. 

\subsection{Models}
\label{sec: TS_models}

We compare a Transformer and Gaussian Process (GP) model against the Martingale property baseline model which assumes that the caseload will not change, i.e. that additional features have zero predictive power. At a forecast horizon $T$ days in the future, the last observed caseload, $\mu_t$, is used to forecast the caseload: $\mu_{t+T} = \mu_t$. 

\textbf{Gaussian Process Model} - 
GP models were shown by \citet{roberts2013gaussian} to perform well in contexts where prior knowledge regarding the appropriate model is limited. The difficulty in inferring the appropriate parametric model in infectious disease modeling led \citet{lampos2017enhancing}, \citet{Lampos2021} and \citet{bin} to adopt a GP time-series model to predict future infectious disease caseloads. More modern methods have since outperformed GP models in time-series forecasting, so this GP model provides a further benchmark to the Transformer model outlined below. Our work uses a radial basis function (RBF) Kernel to specify the covariance function.

\textbf{Transformer model} - 
Transformers have predominantly been used with textual \cite{NIPS2017_3f5ee243} and image-based data \cite{image_transformers}; however, the auto-regressive properties of a masked self-attention layer mean that structurally transformers can obey causality. As a result, many papers have used transformers successfully to model time-series data \cite{lim2021temporal}, \cite{10.1145/3447548.3467401}. Both papers reported that transformer models significantly outperformed the statistical, recurrent and convolutional comparison methods. This success has been replicated in disease modelling by \citet{wu2020deep}. Thus, transformer-based time-series models represent the state-of-the-art in many comparable contexts, motivating their use in this framework. The architecture that is used in this paper mimics that of \citet{NIPS2017_3f5ee243} and \citet{alexandrov2019gluonts}.

\subsection{Time-Series Evaluation}
\label{TS_evaluation}

For the Time-Series task, the prediction error of the forecasts is reported in an ablation study, using the same forecast horizons as the Threshold task. Different feature types make up the covariate set and are compared against the univariate case.

\begin{figure*}[b!]
    \centering
    \includegraphics[width=\linewidth]{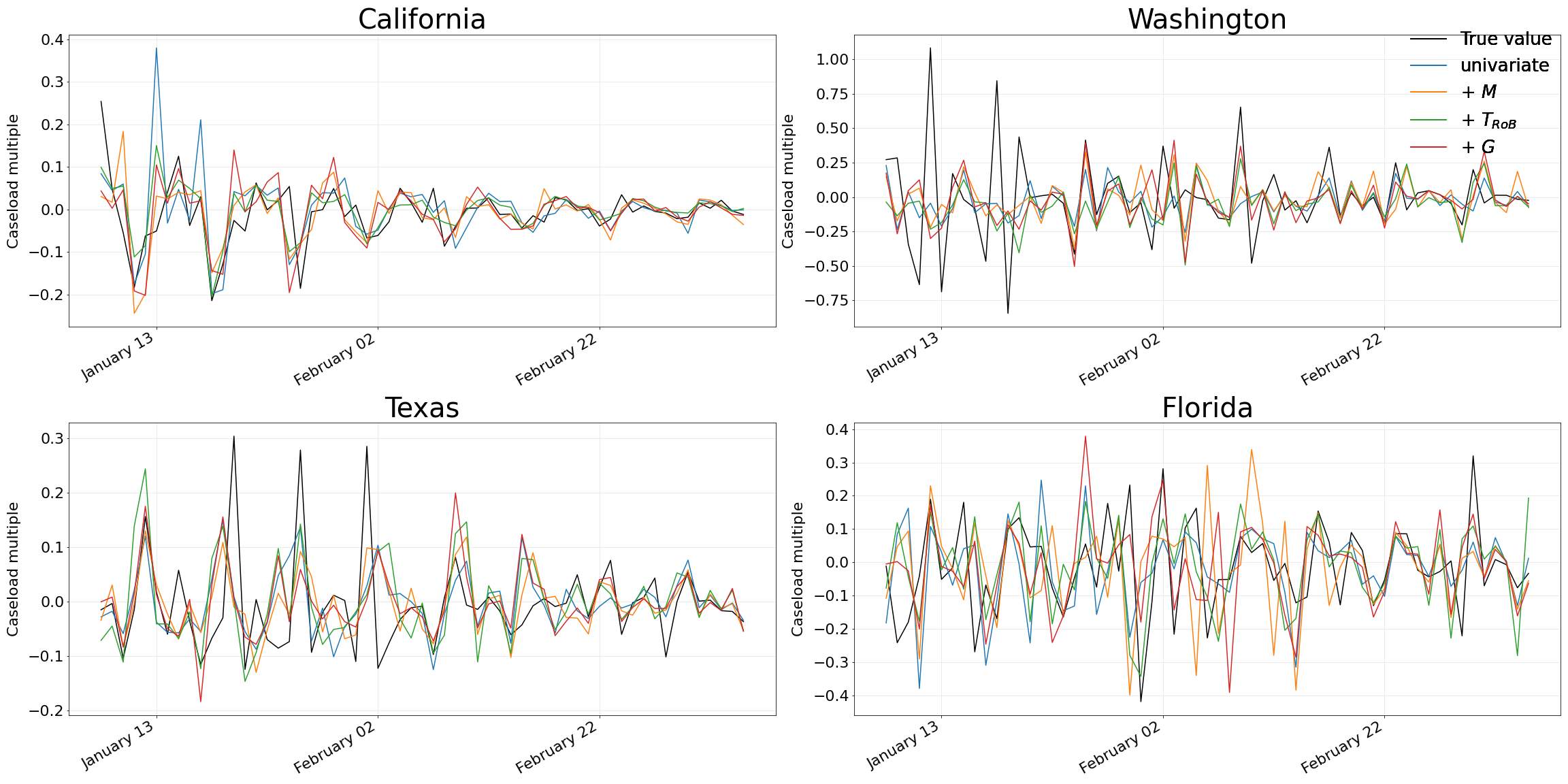}
    \caption{State forecasts at $\tau =7$. The univariate forecast is compared against three multivariate forecasts where the $M$, $G$ \& $T_{RoB}$ features make up the covariate set.}
    \label{fig:forecasts}
\end{figure*}

Table \ref{tab: TS_forecast_ablation} shows the main ablation study, which averages the root-mean-square-error (RMSE) across the different forecast horizons and states. The results show the overall behaviour of the different feature types. The first conclusion is that the Transformer model always outperforms the Martingale and GP model. Poor GP model results are also seen in \citet{Lampos2021}, where their persistence model outperforms the univariate and multivariate GP forecasts in multiple countries. Due to the GP model's weaker performance, further analysis will involve the Transformer model.
State-level results are displayed in Figure \ref{fig:forecasts} and show that the Transformer performs well at modelling the time-series data.

\begin{table}[t!]
    \small
    \centering
    \begin{tabular}{cccccc}
        Data Source & Martingale & GP & Transformer \\
        \thickhline
        $univariate$ & .0366 & .0336 & \textbf{.0291} \\
        \hline
         + $T_{RoB}$ & '' & .0298‡ & \textbf{.0284†} \\
         + $M$ & '' & .0321‡ & \textbf{.0289} \\
         + $G$ & '' & .0308‡ & \textbf{.0290} \\
        \hline
         + $T_{RoB}$ + $G$ & '' & .0322‡ & \textbf{.0288*} \\
         + $T_{RoB}$ + $M$ & '' & .0298‡ & \textbf{.0288*} \\
         + $M$ + $G$ & '' & .0331* & \textbf{.0287*} \\
        \hline
         + $T_{RoB}$ + $G$ + $M$ & '' & .0326† & \textbf{.0288*} \\
    \end{tabular}
    \caption{The RMSE error averaged across all forecast horizons and states. The significance of each result in comparison to the univariate case is denoted by asterisks: * -  P<.2; † -  P<.05; ‡ -  P<.01}
    \label{tab: TS_forecast_ablation}
\end{table}

\begin{table}[b!]
    \scriptsize
    \centering
    \begin{tabular}{ccccc}
        Data Source & Av. & 7D & 14D & 21D \\
        \thickhline
        $uni$ & .0291 & .0274 \cellcolor[gray]{.8} & .0287 & .0301 \\
        \hline
         + $T_{RoB}$ & .0284† \cellcolor[gray]{.6} & .0270* \cellcolor[gray]{.6} & .0274‡ & .0288‡ \cellcolor[gray]{.6} \\
         + $M$ & .0289 & .0274 \cellcolor[gray]{.8} & .0286 & .0292‡ \\
         + $G$ & .0290 & .0274 \cellcolor[gray]{.8} & .0283* & .0300 \\
        \hline
         + $T_{RoB}$ + $G$ & .0288* & .0271* \cellcolor[gray]{.8} & .0287 & .0289‡ \\
         + $T_{RoB}$ + $M$ & .0288* & .0270* \cellcolor[gray]{.8} & .0278‡ & .0298* \\
         + $M$ + $G$ & .0287* & .0271* \cellcolor[gray]{.8} & .0274‡ & .0299 \\
        \hline
         + $T_{RoB}$ + $G$ + $M$ & .0288* & .0273 \cellcolor[gray]{.8} & .0274‡ \cellcolor[gray]{.6} & .0299 \\
    \end{tabular}
    \caption{The RMSE error of a Transformer model averaged across all states at varying forecast horizons, using the same highlighting criteria as Table \ref{tab: time_delay}. The significance notation is: * -  P<.2; † -  P<.05; ‡ -  P<.01}
    \label{tab: TS_forecast_fh}
\end{table}

\begin{table*}[t!]
    \small
    \centering
    \begin{tabular}{c|c|c|cl}
        Size rank & Topic & ID & Frequency & Top 5 words \\
        \hline
        1 & Masks & 95 & 10699 & mask, wear, masks, wearing, gloves \\
        2 & Unemployment & 138 & 7591 & unemployment, claim, pay, money, rent \\
        3 & Appreciation & 181 & 3508 & thank, thanks, appreciate, good, sharing \\
        4 & Schools & 120 & 2808 & school, kids, schools, teachers, students \\
        5 & Temporal statistics & 152 & 1290 & weeks, phase, ago, months, week \\
        6 & Lockdown frustration & 75 & 1217 & closed, shut, f**k, close, die \\
        7 & Agreement & 197 & 892 & yes, agree, yeah, exactly, sure \\
        8 & Festivities & 96 & 879 & thanksgiving, christmas, family, people, party \\
        9 & Vaccines & 196 & 877 & vaccine, vaccines, vaccinated, vaccination, people \\
        14 & Illness & 178 & 569 & cough, fever, symptoms, asthma, throat \\
        17 & Gyms & 50 & 487 & gym, gyms, fitness, open, exercise \\
        19 & Trump & 218 & 378 & trump, people, stupid, inslee, president \\
    \end{tabular}
    \caption{The notable clusters from the r/CoronavirusWA subreddit using a SRoBERTa-NLI-STSb-base language model. The Frequency column represents the number of comments that are included in the cluster.}
    \label{tab: ETM_features}
\end{table*}

The more meaningful conclusion that can be drawn from Table \ref{tab: TS_forecast_ablation} is that whilst the success of the $T_{RoB}$ features is still present, it is more marginal than in the Threshold task both when the $T_{RoB}$ features are used in isolation and when the feature sets are combined. Using the $T_{RoB}$ features in the covariate set does deliver a statistically significant result, however, the decrease in error rate is small. Whilst all feature sets deliver a performance increase, none of the other feature sets can be considered to deliver a statistically significant result. Combining further data types doesn't deliver the expected performance increase, with the performance plateauing, and in some cases decreasing as the number of feature types increases. 
One possibility is that the information that the $T_{RoB}$ features provide is counteracted by the performance costs of having a large number of variables. Table \ref{tab: TS_forecast_fh} reinforces what is seen in Table \ref{tab: TS_forecast_ablation}, showing that across all forecast horizons there is a slight improvement in performance and that using the $T_{RoB}$ features alongside the time-series data generally results in the lowest error across all tested feature sets.  

The significance of the results in Table \ref{tab: TS_forecast_ablation} is calculated by taking 10,000 samples at every forecast horizon for each state. The error of each forecast is calculated, resulting in an error distribution for all feature sets. To discern whether the addition of a feature type results in a statistically significant performance shift a Z-test is used. The univariate forecast is assumed to be the population distribution and each feature set's forecast errors are treated as the sample distribution. The Z-score is calculated using parameters from both distributions:

\begin{equation}
    Z = \frac{\overline{X}_{pop} - \overline{X}_{sample}}{\sqrt{\sigma^2_{{X}_{pop}} + \sigma^2_{{X}_{sample}}}}
    \label{eq:z_stat}
\end{equation}

\begin{table*}[t!]
    \renewcommand\thetable{11}
    \small
    \centering
    \begin{tabular}{c|c|c|cl}
        State & Topic & ID & Importance & Top 5 words \\
        \hline
        \multirow{5}{*}{Washington} & Working & 107 & .21 & work, office, home, headquarters, let \\
         & Illness & 178 & .14 & cough, fever, symptoms, asthma, throat \\
         & Quarantine & 136 & .08 & quarantine, facility, people, outside, think \\
         & Schools & 120 & .08 & school, kids, schools, teachers, students \\
         & Statistics & 150 & .07 & trendline, graph, using, ggplot2, plotted \\
         \hline
         \multirow{5}{*}{California} & Illness & 163 & .12 & cough, throat, fever, chest, symptoms \\
         & School closure & 122 & .09 & schools, close, school, closing, closed \\
         & Guns & 60 & .09 & gun, guns, shoot, firearms, buy \\
         & Safety & 103 & .08 & safe, stay, luck, protect, safer \\
         & Flu & 166 & .06 & flu, pneumonia, influenza, season, spanish \\
         \hline
         \multirow{5}{*}{Texas} & Data & 130 & .20 & source, data, information, info, sources \\
         & Voter Fraud & 72 & .10 & vote, mail, voter, voting, fraud  \\
         & Houston & 78 & .07 & houston, harris, county, area, houstonian \\
         & Doctors & 138 & .06 & doctor, doctors, medical, physician, telemedicine \\
         & Illness & 155 & .04 & fever, cough, allergies, asthma, symptoms \\
         \hline
         \multirow{5}{*}{Florida} & Spring Break & 22 & .22 & spring, break, bike, week, breakers \\
         & Social Media News & 111 & .19 & reddit, facebook, news, echo, chamber \\
         & Statistics & 106 & .05 & numbers, data, believe, trust, graph \\
         & Illness & 94 & .04 & drug, people, fever, virus, sick \\
         & Desantis & 67 & .04 & desantis, care, deathsantis, dbpr, ron \\
    \end{tabular}
    \caption{Important features from each of the key states at $\delta_r = 0.6$ \& $\tau = 7$ days.}
    \label{tab:7day_ETM_diff_states}
\end{table*}

\section{Discussion}
\label{discussion}

Examining the contents of our Reddit features can help us understand the information they provide to the prediction task. We took the top 5 non-stopwords from the cluster to characterise each cluster and manually named them for better comprehension. Table \ref{tab: ETM_features} shows the largest clusters from the SRoBERTa-NLI-STSb-base representation of the r/CoronavirusWA subreddit. These topics identify precise semantic concepts that intuitively provide relevant information for a caseload prediction.

As mentioned, the advantage of the Threshold task is that it provides greater interpretability than the more black-box time-series models. Therefore, the Threshold task is used to understand which features are important to the prediction. Table \ref{tab:7day_ETM_diff_states} shows the weightings of the most important features at $\delta_r = 0.6$ \& $\tau = 7$ days. The cultural differences between the states can be seen via these features, most obviously the \textit{Houston} feature in Texas and the \textit{Desantis} feature in Florida. The \textit{Spring Break} cluster is only seen in Florida, a state that is famed for this holiday tradition and was a large contributor to an increase in non-COVID-19 compliant events that resulted in an increase in cases at the beginning of the pandemic. Equally, the \textit{Guns} and \textit{Safety} features in California likely identify the strong negative reaction from the libertarian community within California to what were the most stringent lockdown restrictions from any of the analysed states. The libertarian trait within California is best characterised by the Prop 22 ballot initiative\footnote{URL: https://vig.cdn.sos.ca.gov/2020/general/pdf/topl-prop22.pdf (accessed: 29/12/2021)} which identifies a political attitude not aligned with strict lockdown measures. Alongside these differences, the \textit{Illness} feature is highly weighted in all states. The use of this feature in all short-term predictions might explain the success of prior work that used static tracking words such as ``Influenza'', ``fever'', ``headache'', etc. \cite{samaras}, \cite{iso-etal-2016-forecasting}; discussion about symptoms is indicative of a rise in cases in all states. It is clear, however, that exclusive use of symptomatic features is not optimal, since other topics besides symptomatic conversation are useful for the prediction.

\section{Conclusion}
\label{conclusion}

Reddit data performs well at discerning different trend signals for COVID-19 caseload increases in the Threshold task. Reddit features alone achieved high accuracy at most threshold increases but were especially strong when identifying whether the caseload was likely to double in the next 14 days, achieving an accuracy of .970. That value is seen in the Time-Series task but the performance benefit is not as stark, especially when the number of features increases. The characteristics of Reddit data make it appealing: it is readily available and updated in real-time, offering the means for monitoring infectious diseases in regions where reported data is unreliable; however global Reddit usage is not constant, and not every area has a subreddit, making our exact methodology hard to scale. As Reddit usage increases and disperses around the world or data from another social media site is adapted to fit within our pipeline, the methods used in this paper will become more scalable. Another notable conclusion is that the predictive information within Reddit data is better extracted by including transformer-based language models in the forecasting pipeline. Language model complexity appears to be linked with performance improvements in the Threshold task.  Strong language models allow us to isolate highly specific features predictive of future caseload increases in an unsupervised setting. 

\section{Future work}

More work can be done on feature selection for the Time-Series task. The value of combining all data types is evident in the Threshold task but that value is not seen in the Time-Series task. Developing models that are able to model a larger number of features more effectively could likely yield some performance gains. On top of this, our methodology relies on using textual data that refers to a specific geographic location. Reddit's structure makes this simple; however, more data is needed to replicate our findings in regions where Reddit take-up is low. Geotagged posts and the geolocation of a user's home region are possible avenues for enlarging the social-media dataset. Finally, the unsupervised methodology outlined in this paper can be adapted to other fields in which a social media derived feature set is used, such as quantitative finance, election and logistics forecasting. 

\section*{Acknowledgements}

The first author was funded by the Economic and Social Research Council of the UK via the Grand Union DTP. This work was also supported in part by a grant from the Engineering and Physical Sciences Research Council (EP/T023333/1). We are also grateful to the Oxford-Man Institute of Quantitative Finance and the Oxford e-Research Centre for their support.

\bibliography{bibliography}
\bibliographystyle{acl_natbib}

\appendix

\begin{table*}[h!t]
    \small
    \centering
    \begin{tabular}{ccccccc}
        Clustering algorithm & k & Average & 7 days & 14 days & 21 days & 28 days \\
        \hline
        \multirow{5}{*}{GMM} & 25 & .659 & .799 \cellcolor[gray]{.8} & .645 & .594 & .596  \\
         & 50 & .691 & .838 \cellcolor[gray]{.8} & .691 & .614 & .622  \\
         & 75 & .698 & .845 \cellcolor[gray]{.8} & .689 & .618 & .639 \\
         & 100 & .702 & .831 \cellcolor[gray]{.8} & .703 & .612 & .662 \\
         & 125 & .714 & .827 \cellcolor[gray]{.8} & .709 & .659 & .664 \\
         & 150 & .716 \cellcolor[gray]{.6}  & .850 \cellcolor[gray]{.8} & .718 & .619 & .676 \\
         \cline{2-7}
         \multirow{5}{*}{KM} & 25 & .706 \cellcolor[gray]{.6} & .883 \cellcolor[gray]{.8} & .685 & .620 & .635 \\
         & 50 & .636 & .681 & .714 \cellcolor[gray]{.8} & .622 & .528 \\
         & 75 & .677 & .781 \cellcolor[gray]{.8} & .769 & .607 & .550 \\
         & 100 & .702 & .787 \cellcolor[gray]{.8} & .678 & .647 & .695 \\
         & 125 & .663 & .757 \cellcolor[gray]{.8} & .754 & .611 & .531 \\
         & 150 & .658 & .768 \cellcolor[gray]{.8}& .687 & .621 & .556 \\
    \end{tabular}
    \caption{The average performances of an RF classification model using KM and GMM clustering across all thresholds at different values of $k$ on the r/CoronavirusWA subreddit. The comment-level SDistilBERT-NLI-STSb-base representations' dimensionality was reduced via UMAP. The \colorbox[gray]{.8}{light grey} indicates the highest performing instance of each model setup. The \colorbox[gray]{.6}{dark grey} indicates the highest average performing model configuration.}
    \label{tab: clust_tuning}
\end{table*}

\begin{table*}[t!]
    \small
    \centering
    \begin{tabular*}{\linewidth}{c @{\extracolsep{\fill}} ccc|cccc}
        Language model & Dim. reduction & Clustering & Average & 7 days & 14 days & 21 days & 28 days \\
        \hline
        \multirow{6}{*}{DistilBERT} & \multirow{3}{*}{PCA} & HDBCSAN & .722 & .821 \cellcolor[gray]{.8} & .724 & .678 & .667 \\
        & & KM & .716 & .807 \cellcolor[gray]{.8} & .675 & .677 & .704 \\
        & & GMM & .714 & .808 \cellcolor[gray]{.8}& .715 & .651 & .680 \\
         \cline{2-8}
        & \multirow{3}{*}{UMAP} & HDBCSAN & .807 \cellcolor[gray]{.6} & .905 \cellcolor[gray]{.6} & .827 \cellcolor[gray]{.6} & .759 \cellcolor[gray]{.6} & .737 \cellcolor[gray]{.6} \\
        & & KM & .706 & .883 \cellcolor[gray]{.8} & .685 & .620 & .635 \\
        & & GMM & .716 & .850 \cellcolor[gray]{.8} & .718 & .619 & .676 \\
          \hline \hline
         & & Average & .730 & .846 \cellcolor[gray]{.8} & .724 & .667 & .683 \\
    \end{tabular*}
    \caption{The average performance, on the r/CoronavirusWA subreddit, of an RF model across all thresholds at different prediction horizons for each of the model pipelines using only $T_{RoB}$ features. The variables and highlights are the same as in Table \ref{tab: clust_tuning}.}
    \label{tab: dim_red+cluster test2}
\end{table*}

\section{Clustering algorithms hyperparameter tuning}
\label{app: clust_tuning}

\begin{figure}[h!]
    \centering
    \includegraphics[width=\linewidth]{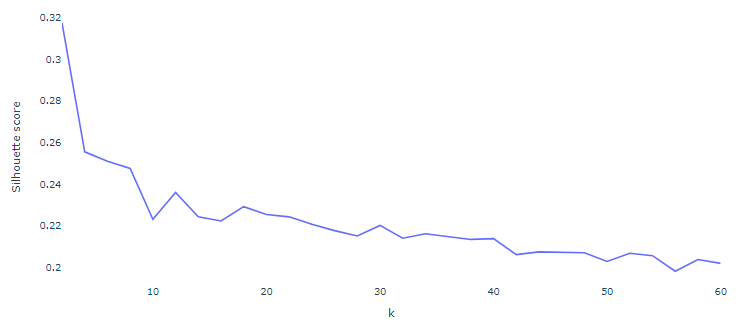}
    \caption{Silhouette score vs. $k$ using a KM clustering algorithm on UMAP-DistilBERT embedding space.}
    \label{fig:silhouette_score}
\end{figure}

An exhaustive search has been conducted to find the optimal $k$ parameter (number of clusters) for KM and GMM clustering to compare their optimal configurations against HDBSCAN. The standard Silhouette score method was trialled for fine-tuning the $k$ parameter, but the result was $k=1$, perhaps indicating the unsuitability of KM and GMM for this task. Figure \ref{fig:silhouette_score} is a plot of the Silhouette score \citep{silhouette} of a KM clustering algorithm for different values of $k$ on the UMAP reduced SDistilBERT-NLI-STSb-base embeddings space. 
The maximum Silhouette score should be the most appropriate k value if the data is divided into distinct clusters. The maximum Silhouette score at $k=1$ indicates that the data is structured into one central cluster with high and low-density areas. 

Since the Silhouette score does not provide an obvious $k$ parameter, and yet there needs to be some proof that HDBSCAN is a better algorithm than KM and GMM, an exhaustive search for the optimal $k$ on the development data is conducted to prove that KM and GMM are not suitable for the task. $k$ is tuned using the performance on the r/CoronavirusWA data from 01/03/2021 to 17/01/2021 with UMAP dimensionality reduction. 

The same search was conducted to find the optimal $k$ for the PCA space; for KM, the value was 75, and for GMM, the value was 125 on the DistilBERT embedding space. These values of $k$ were used for the testing in Appendix \ref{app: ETM_dim_clust}.

\section{Dimensionality reduction and clustering algorithms}
\label{app: ETM_dim_clust}

A test was carried out to see which combination of dimensionality reduction and clustering algorithms resulted in the best overall performance. The different algorithms were tested using SDistilBERT-NLI-STSb-base representations of the comments. The two dimensionality reduction techniques used were PCA and UMAP; the three clustering techniques used were GMM, KM and HDBSCAN. The $k$ values derived in Appendix \ref{app: clust_tuning} were used for the GMM and KM clustering, and the evaluation pipeline used is the Threshold task described in Section \ref{sec: thresh_class}.

The results from Table \ref{tab: dim_red+cluster test2} show that the combination of UMAP and HDBSCAN is the best combination of algorithms; the UMAP-HDBSCAN combination is the best performing pipeline across all prediction horizons.

\section{Aggregated Sequence Classification models}
\label{app: ASC_models}

As mentioned in Section \ref{introduction}, the most obvious way to incorporate the modern transformer-base language models is to formulate the problem as an Aggregated Sequence Classification (ASC) task. It has been shown that BERT and other similar models are well adapted to performing sequence classification, and this has become a common usage of these language models \cite{devlin-etal-2019-bert}. Therefore, it is important to trial a model that incorporates this more standard methodology before trialling other feature identification methods. 

For evaluation, we trialled two language models: BERT-base-uncased, and a domain adapted version of BERT-base-uncased trained on the r/Coronavirus subreddit - CoFReBERT (\textbf{Co}VID-19 \textbf{F}orecasting from \textbf{Re}ddit \textbf{BERT}). The language models are then fine-tuned on a Sequence Classification task in which the [CLS] token encodes the "up" or "down" class, indicating a possible increase or decrease in the number of cases. The adapted models are referred to as ASC-BERT and ASC-CoFReBERT. The model is trained on balanced classes with a 4:1 train-test split, where each day is assigned to be a test or a train day, and all comments written on a particular day are categorised together. Once the model labels each comment within the test set as either "up" or "down", the majority class on a given test day is assigned as the prediction for that day. 

\begin{table}[h]
    \renewcommand\thetable{10}
    \small
    \centering
    \begin{tabular}{cccccc}
        Models & Av. & 7D & 14D & 21D & 28D \\
        \hline
        ASC-BERT & .631 & .769 \cellcolor[gray]{.8} & .655 & .561 & .537 \\
        ASC-CoFReBERT & .701 & .846 \cellcolor[gray]{.8} & .690 & .634 & .634 \\
        \cline{2-6}
        $T_{RoB}$ & .869 \cellcolor[gray]{.6} & .923 \cellcolor[gray]{.6} & .896 \cellcolor[gray]{.6} & .810 \cellcolor[gray]{.6} & .855 \cellcolor[gray]{.6} \\
        $T_{BoW}$ & .765 & .780  & .808 \cellcolor[gray]{.8} & .804 & .791 \\
        \hline
        \end{tabular}
    \caption{The average performance, on the r/CoronavirusWA subreddit, across all thresholds at four different prediction horizons. The variables and highlights are the same as in Table \ref{tab: clust_tuning}.}
    \label{tab: comparison_models}
    
\end{table}

From the results in Table \ref{tab: comparison_models}, it is clear that the models do not perform as well as hoped in comparison to the traditional static word features and the features outlined in this paper. The main reason for this is likely to be noise from the unsupervised labelling process. Comments that are either unrelated to the prediction or indicate an opposite caseload trend are included in the prediction. Without manual labelling, it is hard to reduce this noise; however, that would result in investigator bias entering the prediction. Furthermore, it is not completely clear whether a comment is indicative of a rise in cases, shown by the variety of topics considered important to the prediction in Table \ref{tab:7day_ETM_diff_states}. Therefore, the structure of the ASC models is not well adapted to the task of predicting COVID-19 cases. 

\section{Training and software details}
\label{app: TS_training}

\textbf{Python Packages} \ The sentence-embedding models from \cite{Reimers2019} were used to encode the Reddit post representations using the \texttt{sentence-transformers} Python package. The time-series models were both implemented using the \texttt{gluonts} Python package \cite{alexandrov2019gluonts}. The ASC models outlined in Appendix \ref{app: ASC_models} use the BERT-base-uncased model from the \texttt{transformers} package and the ASC-CoFReBERT model was trained using the \texttt{run\_mlm.py} file in the library.
\\
\\
\textbf{Training Pararmeters} \ Besides the analysis detailed earlier in the Appendix, we do not perform hyperparameter tuning but use common hyperparameter values for all calculations in this paper. For the Random Forest model in the Threshold task, the number of trees is 100, and the maximum tree depth is 20. The Time-Series models were trained over 50 epochs and used default parameter values. The ASC-CoFReBERT model was trained with standard parameter values, using a batch size of 128 and a dropout probability of 0.1. 
\\
\\
\textbf{Computation} \ All experiments in the main body of the paper were run on a personal computer, the ASC model in Appendix \ref{app: ASC_models} was run on the. The ASC model was run on a Tesla P100 and took between 3 to 6 hours to run, depending on the size of the subreddit.
\\
\\
\textbf{Licenses} \ There are licenses associated with the use of some of the data and Python packages used in this paper. The OxCGRT dataset and Pushshift API are open access under the Creative Commons Attribution CC BY and 4.0 International standards. The COVID-19 Tracking Project, \texttt{gluonts}, \texttt{transformers} and \texttt{sentence-transformers} Python packages are licensed under the Apache License 2.0.

\end{document}